\documentclass[sigconf,screen]{acmart}
\AtBeginDocument{%
  }

\setcopyright{acmlicensed}
\copyrightyear{2024}
\acmYear{2024}
\acmDOI{XXXXXXX.XXXXXXX}

\acmConference[PAI/24]{‹Programming› 2024 Conference Companion}{March 11 -- 15,
  2024}{Lund, Sweden}
\acmISBN{978-1-4503-XXXX-X/18/06}

\usepackage{cleveref}
\usepackage{listings}

\usepackage[utf8]{inputenc}

\usepackage[frozencache,cachedir=.]{minted}
\DeclareUnicodeCharacter{2605}{\star}
\usepackage{xcolor} 

\begin{document}

\title{Magic Markup: Maintaining Document-External Markup with an LLM}

\author{Edward Misback}
\email{misback@cs.washington.edu}
\orcid{0009-0003-9474-0826}  
\affiliation{%
  \institution{University of Washington}
  \city{Seattle}
  \country{USA}}
\author{Zachary Tatlock}
\email{ztatlock@cs.washington.edu}
\orcid{0000-0002-4731-0124}
\affiliation{%
  \institution{University of Washington}
  \city{Seattle}
  \country{USA}}
\author{Steven L. Tanimoto}
\email{tanimoto@cs.washington.edu}
\orcid{0000-0002-8175-7456}
\affiliation{%
  \institution{University of Washington}
  \city{Seattle}
  \country{USA}}

\renewcommand{\shortauthors}{Misback et al.}

\begin{abstract}
    Text documents, including programs, typically have human-readable semantic structure. Historically, programmatic access to these semantics has required explicit in-document tagging. Especially in systems where the text has an execution semantics, this means it is an opt-in feature that is hard to support properly. Today, language models offer a new method: metadata can be bound to entities in changing text using a model's human-like understanding of semantics, with no requirements on the document structure. This method expands the applications of document annotation, a fundamental operation in program writing, debugging, maintenance, and presentation. We contribute a system that employs an intelligent agent to re-tag modified programs, enabling rich annotations to automatically follow code as it evolves. We also contribute a formal problem definition, an empirical synthetic benchmark suite, and our benchmark generator. Our system achieves an accuracy of 90\% on our benchmarks and can replace a document's tags in parallel at a rate of 5 seconds per tag. While there remains significant room for improvement, we find performance reliable enough to justify further exploration of applications.
\end{abstract}



\received{29 February 2024}
\received[revised]{5 March 2024}
\received[accepted]{5 March 2024}

\maketitle

\section{Introduction}

\subsection{Problem statement}

Document markup allows a number of powerful behaviors related to \textit{tagging} text with metadata. As an example of the impact of markup, Hypertext Markup Language (HTML) is famously the backbone of the World Wide Web, which is formed by hyperlink annotations that tag some piece of text in one document, called \textit{anchor text}, with a link to a related document\cite{bernerslee1990}. A curious reader might wonder: has markup led to changes of a similar scale in the realm of software engineering?

In software engineering, code comments are a simple way of marking a document with miscellaneous helpful information. More extended systems for attaching data to code also exist--systems in the domain of \textit{literate programming} integrate information about code with the text of the code itself, and rely on markup to accomplish this\cite{Knuth1984}. Perhaps the JSON files backing Jupyter notebooks can also be considered a form of markup document, if the code that runs is taken as ``the document.''

However, if we think of markup as a basic primitive for working with and referring to text, and think further about the universality of text as an interface in programming, we should be surprised to find that this primitive is almost universally unsupported, even in advanced live programming environments. For code in particular, there is no standard for attaching metadata like code review history and example data to a particular point in the text. We attempt to explain why below.

A principal challenge in designing a document markup system is maintaining the correct positions of text tags when the underlying content that the markup refers to is edited. This is called \textit{annotation anchoring}. The simplest solution to this problem is to include the tags in the document text itself, as in HTML or standard code comments. This requires no special programming tools, but it burdens the document's reader (whether human or computer) with distinguishing content from metadata, so it isn't suitable for documents with many layers of extensive annotation--imagine a line of code with comments left by 10 different people for completely different purposes. The second-simplest solution to this problem is to manage document edits through a special program like a \textit{"What You See Is What You Get" (WYSIWYG) editor} that shows only the document content (with the effects of markup metadata) while managing the positions of tags behind the scenes. The Microsoft Word document system is an example of this second solution, and also shows its downsides: the cost of building and maintaining a special editor that acts the way a human expects is significant, and it also locks the programmer into always using a particular editor. Very few editors can operate on Microsoft Word documents.

This paper advances a third solution to the problem. Prior authors have considered the idea of attaching \textit{external annotations} to programming systems with \textit{text anchoring}--methods that relate metadata kept in a separate file to a point in the code using text similarity. These systems are useful, but fail after the text has changed enough (for example, when the names of variables in a program have changed, or when a loop has been vectorized), even when a human could still find a reasonable new position for the annotation using their understanding of the text's syntax and semantics. As such, serious infrastructure to support ubiquitous external annotations has not been feasible. 

This third solution is the only fully general solution for programming systems and other systems with restrictions on the structure of text in the file. For example, many configuration files are stored as JSON, which simply does not support comments at all in its formal standard. Further, all files presented through a plain text editor have implicit restrictions based on what can reasonably fit in the editor's viewable buffer.

To address this issue, we propose \textit{magic markup}: markup maintained separately from the document by a semantics-aware system that ``magically'' handles re-tagging after document updates. What would it mean to be able to keep markup off of a document, and what would it mean to be able to mark up code?

\subsection{User story}\label{user-story}

To illustrate how we imagine external annotations being used, we present a user story involving two programmers.

Barbara is a senior data scientist who has received an informal code review request from Alex, a junior developer in the same company. Alex has updated the code of a particular function in the production code base responsible for an image classification task. This code includes an image transformation pipeline with subparts that are known to be performance-intensive.


Barbara begins by pulling Alex's changes and opening the updated file in her editor. The code review request appears in Barbara's editor as a set of annotations next to the file. Alex intends these annotations as review requests for Barbara only. These are \textit{user-directed comments}--other programmers using the code base may see these comments, but since they are stored in a separate database (that was pulled with the code), they are not part of the file and hidden by default for everyone but Barbara. As Barbara begins to edit the document to address issues she sees in the changes, the comments remain attached to the entities she would expect. Even when she vectorizes a loop that Alex mentioned he was uncertain about, completely changing its text, the document's tag maintainer---an intelligent agent backed by a language model---knows the vectorized code is intended to replace the loop, and the comment remains in the right place until she marks the concern as completed. She knows Alex will easily locate her update through the resolved comment, even though she has also moved the vectorized code into a separate function.

She notices that Alex added a new nested lambda function that introduces additional image processing. She isn't familiar with the method Alex used, but fortunately, Alex annotated this part of the pipeline with some \textit{example data}, and she looks at the cached output image \textit{visualization} annotation for a moment before it updates with the new output from the \textit{out-of-context execution} of her own system on just the lambda function with Alex's example data. Barbara remembers the days when she would have had to copy this code out into a REPL or a notebook and synthesize example data herself just to check its behavior and throw it all away afterward. The behavior of this section is still a little unclear for quick reading, though, so she selects the section and asks a language model to generate a short clear \textit{dynamic explanation} of its function that will continue to apply as the code base evolves.

Barbara looks at the next part of the new code--a loop body with a performance concern. She decides that she will have to check on this part herself, so she selects the section and asks her editor to run just that section with the output from Alex's section and time its performance across 5 executions. The performance is not as good as it should be, and she realizes Alex's output is unnecessarily high-resolution. She quickly fixes this, and in response, the output image updates and the execution time drops. She asks her editor to \textit{warn} the programmer if this example execution time ever goes above 20 milliseconds, as that probably would have helped Alex. This annotation automatically becomes part of the document's \textit{performance overlay}, which is different from the \textit{presentation overlay} she uses when walking new developers through the code. Again, she remembers how troublesome it would have previously been to unit test this loop body.

Barbara also notices an edit to a data structure that she realizes it would be best to mark as off-limits to the junior developers on the team. She adds a note that all of the junior developers will see when opening that section of the code.

Alex's change introduces a new flag for the image processor, and Barbara follows a new \textit{documentation link} on the flag to the relevant documentation section Alex added for this. The documentation links back to particular code blocks Alex added when referencing implementation details.

Having a last quick look through the file, Barbara finally fixes the first minor issue she noticed--there was a typo that broke parsing at the top of the file. \textit{This didn't matter for any of her other interactions with the file}, since it wasn't in the annotated sections and didn't affect the segment-specific executions.

All of Barbara's notes and tests from the review can be consulted by any team member later, with the new annotations laid out on a timeline that reconstructs Barbara's thought process as she worked with the code.

Even if someone edits the file in an unsupported editor, Barbara's team trusts the tag maintainer to correctly re-tag the document afterward without any issues, flagging any seriously ambiguous re-taggings for their review. The platform-independence of the tag system lets them forget about the tags when they don't need the extra information, and even lets them maintain annotations on the source code of an independent code base for one of their dependencies that they only have read access for.

Barbara's rapid, high-level code review is the product of tools whose foundation is a highly reliable tag maintenance system with the following properties:
\begin{itemize}
    \item Annotation anchors are updated each time the code is modified.
    \item Annotations are stored separately from the base document to avoid breaking editing and execution for typical text editors and program interpreters and to keep the annotations of different tools independent.
\end{itemize}

\subsection{Our contributions}
The above vision leads us to seek answers to the following questions:

\begin{enumerate}
    \item How capable are current language models as ``tag maintainers'' for a document? Are they reliable, fast, and cheap enough to build on top of?
    \item What kinds of documents or edits make tags hard to maintain? For source code in particular, what kinds of edits exist at the semantic level, and in what cases does simply moving tags or noting they have been ``orphaned'' fail to capture that meaning?
\end{enumerate}

To answer these questions, we construct an LLM-based re-tagging system. To evaluate our re-tagging system and promote further progress on this problem, we synthesize a benchmark suite representing 90 code updates across 5 programming languages in which a tagged entity is relocated or altered. We also provide the code for benchmark generation.

Our contributions include the following:

\begin{enumerate}
    \item a formal vocabulary for the problem space that introduces the notion of annotation intent
    \item adaptable code for generating empirical test and training data for the re-tagging task
    \item the synthetic Tagged Code Updates benchmark dataset, generated with the above and cleaned
    \item an LLM-based re-tagging system
    \item an evaluation of our system's performance on the benchmarks using OpenAI's GPT-4 Turbo\footnote{gpt-4-0125-preview} model
\end{enumerate}







\section{Related work}

The problem of anchoring text has a long history. In systems that are able to observe edit actions, schemes like the "sticky pointers" of Fischer and Ladner \cite{fischer1979data} can be used, but offline systems need methods for dealing with arbitrary document updates. Brush et al. describe work on robust annotation systems for digital documents (like Microsoft Word documents) that account for user expectations using methods like keyword anchoring \cite{brush2001robust}. They note that annotation orphaning (loss of tag position) is a key problem in these systems, and that systems typically have strategies for dealing with orphans.

Prior work has also investigated the use of annotations for source code. Juhár \cite{juhar2019supporting} distinguishes language-level annotations that are part of code, structured comment annotations, and external annotations requiring the support of a special system, typically an IDE. These types differ in how annotations are defined, applied, and in what code elements they are able to bind to, and Juhár develops an IDE-based annotator that maintains annotation positions externally.

Keeping annotations attached to the right entities is closely related to the question of how to track changes in code over time. Reiss labeled changes in 53 lines of code across 25 versions of a Java source file, then evaluated 18 tracking methods with various parameters to find that a relatively simple, low-computational-cost combination of string similarity \cite{levenshtein1966binary} and context comparison yielded the best results \cite{reiss2008tracking}. Reiss's tracking method has been used in other systems for attaching annotations to a tracked line, including Horvath et al.'s Catseye \cite{horvath2022using} and Sodalite \cite{horvath2023support} systems for adding comments to source code and allowing users to reference source code in documentation, respectively. Horvath et al.'s idea in these systems is notable because it is very similar to our own: when Catseye or Sodalite load, they attempt to reattach old annotations to the file using Reiss's tracking method. Horvath et al. note that this method was able to resolve 86.5\% of cases in their testing. However, the method has no semantic awareness and still suffers from the text similarity issues we've noted previously.

Our key insight follows historical discussions about notions of program similarity. Walenstein et al. break program similarity into two types--representational similarity (concerning text, syntax, and structure) and semantic or behavior similarity (concerning a program's function or execution) \cite{walenstein2007similarity}. We are interested in the potential for language models to address the latter form, and our solution handles that prior string similarity-based methods cannot by taking advantage of this capability.

\section{Basic Definitions}\label{sec:basic-definitions}

While the remainder of this paper presents an early exploration into
the power of LLMs to update annotations in the context of evolving
code bases, this and future work can benefit from a clarification
of the terms and concepts involved in this research.  This section
both addresses this need and suggests a longer-term trajectory
of work that takes account of explicit notions of the intent of
annotations while maintaining them.

Due to the wide variety of uses for annotations in documents
and the implications of usage contexts for automatic maintenance
of annotations during document editing, we propose terminology
to clarify some of the otherwise ambiguous notions on this topic.
We start with the simplest concept of ``text point'' and work
through ``annotation" and finally ``mapped annotation''.

A {\em text point} $TP$ is the character index (an integer, zero-indexed) used to designate
a position in some (any) text string. The text point is independent
of any text, except to the extent that the text be long enough to
have position corresponding to the text point. For example, the
text point 4 refers to the position of ``C'' in ``ABRACADABRA'' and to
the position of the second ``d'' in ``Aladdin'' but does not refer to
anything meaningful in ``Fun''.  We'll say that text point 4 is
compatible with ``ABRACADABRA'' and ``Aladdin'' but incompatible with 
``Fun''.

A {\em text segment} $S$ of a document $D$ is a part of $D$ specified by
a starting text point $TPstart$ and an ending text point $TPend$, where both
text points are compatible with $D$. The substring of $D$ that
starts at $TPStart$ and ends right before $TPend$ is considered part of $S$.
Thus $S$ = [1, 4, ``lad''] is a text segment of ``Aladdin'' but
not of ``ABRACADABRA'' and not even of ``The boy Aladdin''.

An {\em annotation} $A$ of a document $D$ consists of a text segment $S$ together
with two additional pieces of information:\\
  (i) contents.  We can assume this is text or hypermedia represented
textually (e.g., with HTML).\\
  (ii) intent.  Though often unknown or unspecified, this is a kind of
metadata associated with the contents and the text segment that can be
important in the accurate maintenance of the annotation as the underlying
document $D$ goes through edits or other transformations.

The text segment $S$ of annotation $A$ is known as its {\em anchor} \cite{brush2001robust}.
The substring of $D$ in $S$ is known as the {\em anchor text} of $A$.
If the substring is of length 0, then the anchor is called a {\em point anchor}.
Otherwise, it is called a {\em range anchor}.

A {\em document view} $V$ consists of a document $D$ and a set $Z$ of annotations.

Given a document view $V = (D, Z)$ and a transformed (e.g., edited) version
of the document $D'$, the view-mapping problem is to update $Z$ to obtain that
$Z'$ which best respects the intents of annotations in $Z$. We call $Z'$ the
mapped annotations from $V$, and $V'$ is the mapped view of $V$.

As an example, let $D$ = ``boy alad.'' Let $Z = \{A_1\}$ where $A_1$=([4 , 8, ``alad''],
content:``The boy's name'', intent:`TRACK NAMES').  Further let
$D'$ = ``The young boy Aladdin wandered out.''  We may expect that
$Z' = \{A_1'\}$ where $A_1'$ = ([14, 21, ``Aladdin''],
content:``The boy's name'', intent:`TRACK NAMES').

The incorporation of intent as a component of an annotation is not
customary in computer technology at this time.  For example, highlighting
tools in editors do not require indication of intent or directly offer any
specific affordance for expressing the intent of an annotation.  Microsoft Word and Powerpoint do offer
alternatives to highlighting: strikeout, underline, change of font, font size, or
color, etc.  But these do not enforce any clear intent communication either.

However, intent is actually a fundamental aspect of annotations as defined outside
of the computing-tool context.  Here is the Merriam-Webster definition:
\begin{quotation}
``annotation: (noun) a note added by way of comment or explanation.''
\end{quotation}
To unpack this definition, we can identify three essential aspects of an
annotation.  First, the ``note'' aspect is its content, some text or similar
representation of the annotator's idea.  The word ``added'' indicates that
the note is a new component of something existing -- that would be the
base document or the base document already joined with other annotations.
Finally, there is something about the purpose of this added note:
``by way of comment or explanation''. The phrase ``by way of'' can be
interpreted here as ``based on'' or ``associated with'', and the objects
that might be associated are ``comment'' and ``explanation''.  These can be
considered purposes, motivations, or reasons for the annotation.  They are
communicative modes or roles for the annotation.  They are examples of what
we are naming, in this paper, ``intents''.

Intents of annotations are very important inputs to any process of maintaining
annotations when the underlying document for it changes.  Without any explicit
information about intents, any algorithm will have to guess or embody a
programmer's guess about intent of annotations in a given application.
A typical yellow-marker highlight in a text is an annotation with a range
anchor and anchor text, but no content.  A good guess at the intent in such
an annotation is to express ``I think this anchor text is relatively important in this
document.''   This may be enough to guide a smart annotation mapper to a good
result, because it implies that there is some semantic integrity of the anchor
text that should be preserved in the mapped annotation, and the location (of the
anchor text segment) should be updated in such as way as to maintain the same
relationship between the segment and the surrounding context before and after
the transformations.

A very different intent of an annotation is the marking of line numbers, which
although not present in the original document, can support an editorial process
or programmer interaction with a debugger.  When the program is edited, we do
not expect that the line numbers will follow the syntactic or lexical contexts in which they
originally occurred, but will reflect the new line structure post-editing.  Another example of an
annotation that should not be updated according
to surrounding context is a point-anchored annotation with content ``5000 words to here.''
Without either explicit intent or correct inference or guessing of intent, an automatic
annotation mapper would be at a loss to do the best thing.

\begin{figure*}
    \centering
    \includegraphics[width=1\linewidth]{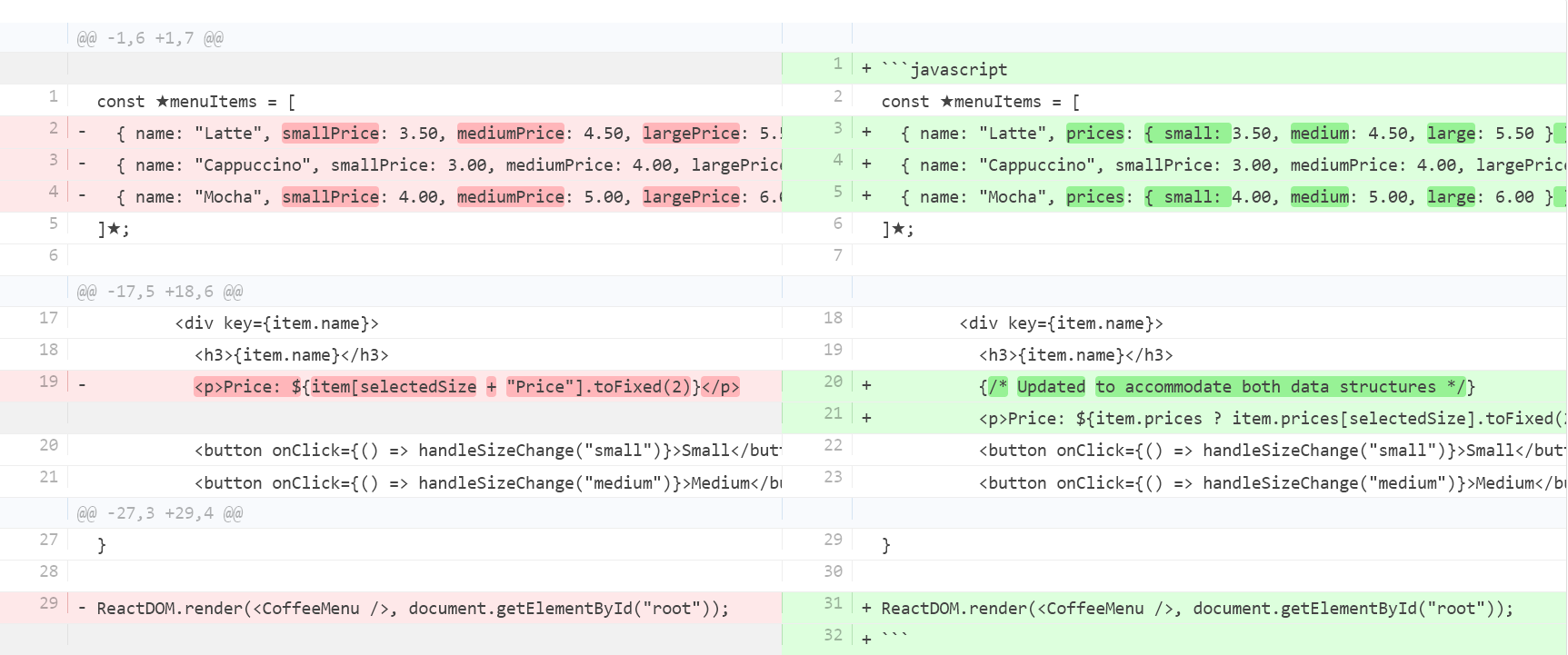}
    \caption{An example synthetic benchmark. On the left, a language model has produced an original program for displaying the price of drinks, and another model has selected and delimited a segment (the ``menuItems'' constant) with black Unicode star characters ($\star$). On the right, a language model has synthesized updates to the program while keeping the segment in place. Our re-tagging system predicts the position of the segment in an \textit{unmarked} version of the file on the right.}
    \label{fig:program-update}
\end{figure*}

\section{Tagged Code Updates Benchmark Suite}

To define the targets for a re-tagging system empirically, benchmarks are needed. While real-world codebases like the Java files tracked by Reiss et al. \cite{reiss2008tracking} should be the gold standard for such benchmarks, language models present an opportunity to rapidly construct synthetic benchmarks with significant detail about the intent behind refactorings. We create the Tagged Code Updates benchmark suite as an example.

Our benchmark suite and generation system are available on \href{https://observablehq.com/@elmisback/tagged-code-updates-benchmark}{Observable} and \href{https://github.com/elmisback/magic-markup/tree/main/benchmark}{Github}.

\subsection{Code generation system}
Our system generates examples of code that has undergone a single edit or refactoring (Figure \ref{fig:program-update}). Examples are generated through a series of queries to language models of varying capability. We note that steps involving creative generation can be performed by any moderately creative and attentive model, but we required one of the largest available models (GPT-4 Turbo) to reliably perform steps involving correctness.

For each example, we begin with this prompt:
\begin{quotation}
Briefly describe an intermediate-level \$LANGUAGE programming problem including at least one \$SNIPPET\_TYPE that can be solved in a single file. Use a creative, real-world framing. Describe steps to solve this problem. Do not provide code yet.
\end{quotation}
This step creates a high-level frame for the code that will be produced. A full example output can be seen in the appendix (section \ref{problemDescriptionExample}).

Next, we generate initial code to solve the described problem. We found that this typically results in small, easily-understood programs that might be used as applications examples in beginner and end-user programming contexts, like a simple calculator for a pizza restaurant or a grocery store inventory manager; perhaps the most interesting program we saw was a genetic algorithm for optimizing the delivery routes of an e-commerce driver (benchmark 64).

After this, a short delimiter string that the model will not confuse with the text already in the file is chosen to delineate the sections of code that are being annotated. We hardcoded this value as a Unicode star character (U+2605, 
$★$) for our benchmarks, since it did not occur in any of the generated code. An extended system might have to dynamically handle a variety of possible character sets.

At this point, the model is asked to describe a snippet in the program matching the SNIPPET\_TYPE. This again creates a high-level frame for the next step, where the model is asked to rewrite the code with the described snippet delimited using the chosen delimiter string. Adding the delimiters was the first time we found it helpful to use the largest available model, in order to make sure the language of the snippet description and the actually delimited segment matched our expectation as programmers as closely as possible. (In retrospect, we find the snippet descriptions in our benchmarks still vague. Generating the snippet description itself with a larger model and more detailed instructions about connecting the snippet with the code would probably create a closer match.) An example of a generated snippet can be seen in the appendix (section \ref{snippetDescriptionExample}).

Next, we ask the model to describe ``an interesting change or refactoring of this code that a real-world programmer might apply.'' We provide a flag that optionally asks the model to ``[d]escribe a state where this code change has only been partially applied'' to try to obtain more realistic examples of incomplete edits or refactorings. An example of an update description can be seen in the appendix (section \ref{updateDescriptionExample}).

Finally, we generate the updated code, specifying that the delimiter position must be preserved, with their contents ``functionally identical in the new version of the code.'' Figure \ref{fig:program-update} shows an example output benchmark; see the appendix for the full prompt (section \ref{updatePrompt}). 

\subsection{Benchmark suite description}

\subsubsection{Parameters}

We initially generated 101 program/program update pairs for our benchmarks. 
The suite targets 5 languages: Python, Javascript, JSX, Racket, and C.
We wondered if different languages and syntaxes 
might present different levels of difficulty for the model's re-tagging attempts.

We asked the model to choose from 6 kinds of snippets: 
constants, subexpressions, variable assignments, 
loop bodies or code blocks, loop conditions, and function calls.
These represent a slightly-diverse sample of common simple code structures, but leave out many other possibilities, like selecting keywords, parts of comments, or multiline sections of code.

We also generated an additional 10 examples for a ``training set'' used for prompt tuning, described in section \ref{tuning}.

\subsubsection{Benchmarks filtered out of the test set}
\label{eval:filtered}

After generation, we manually reviewed all benchmarks and excluded 11 from the final set. 8 of these were excluded for generation mistakes like missing delimiters in the initial or output code.\footnote{We did later test those with proper initial delimiters separately from our evaluation below to see if our system had any problem with re-tagging these examples, but it did not.} We also found 2 other kinds of outputs:
\begin{enumerate}
    \item The segment is completely missing in the updated code. For example, a loop condition is gone after the loop has been refactored into a reduce() call. (2 benchmarks)
    \item Multiple independent segments could be delimited due to ambiguity after code duplication (1 benchmark).
\end{enumerate}

2 and 3 represent more complicated cases that we decided were out of scope for this benchmark suite. Other benchmarks designed to target those particular issues are needed to explore them properly.

This left us with a test set of 90 examples, with 28 written in Python, 17 in Javascript, 17 in Racket, 16 in JSX, and 12 in C. The types for the targeted snippets were as follows: constants (17), subexpressions (19), variable assignments (12), 
loop bodies or code blocks (12), loop conditions (13), and function calls (17).

\subsubsection{Benchmark suite generation costs}
\label{eval:cost}

Each benchmark required around 1800 input tokens and 700 output tokens from a ``large'' model like GPT-4 and 1300 input tokens and 1000 output tokens from a ``small'' model like GPT-3.5. This cost us roughly \$.04 per benchmark through OpenAI's platform. Generation time for each benchmark was on the order of tens of seconds due to the relatively large number of output tokens required from a large model.

\section{Prototype re-tagging system}

We created a re-tagging system\footnote{Available on \href{https://github.com/elmisback/magic-markup}{Github} and \href{https://observablehq.com/@elmisback/magic-markup-retag}{Observable}.} to measure the capability of current language models on the Tagged Code Updates benchmark suite. Our system submits a single prompt to obtain the text and line numbers in the file of the updated segment, then matches the text points for the beginning and ending of the segment in the updated file.

\subsection{Re-tagging prompt}
An example of an application of our re-tagging prompt template can be seen in the appendix (section \ref{retag-prompt}). To allow the model to reference line numbers reliably, we prepend the line numbers for the original and updated files. In case the tag's contents are repeated in the section indicated by the line numbers, for example in a single line like $a = ★a★ + a$, we also request the index of the correct match's occurrence in the section.
We use the OpenAI JSON response format to constrain the model to only generate valid JSON in its response.

\subsubsection{Prompt hand-tuning}\label{tuning}

Before settling on this prompt for our evaluation, we used the 10 examples in our training set to check variations, including variations that were more successful for less powerful models (gpt-3.5-turbo-0125, Gemini Pro, and Mistral8x7b).\footnote{Although it was trained on code in particular, CodeLlama-70b-Instruct-hf did not follow instructions well enough to be considered.} These variations included:

\begin{itemize}
    \item Prompts that simply generate the full re-tagged file. These prompts were initially promising, especially when run on a large model, but we found that the output was not faithful enough to the text of the updated file to trust. For example, the output might be missing comments, or have whitespace differences, or a single typo somewhere in a long file. Furthermore, the time required for a large model to copy an entire file is significant.
    \item Prompts that generate plain English responses. This sometimes helped smaller models in our experience. However, parsing the generated output introduces an additional point of failure.
    \item Prompts that first ask the model to focus on a smaller section of the file. This significantly helped smaller models in our experience. In particular, asking the model to reprint the general section of the file it was focusing on seemed to help it locate finer-grained segments of that section.
    \item Prompts that ask the model to reevaluate previous answers. This did not seem to help smaller models when the previous answer was ``known'' to be the model's own output. A proper setup might ask the same prompt several times and try to average or take the best idea across the results.
    \item Prompts that ask the model for a confidence rating or ask the model to ``put yourself in another programmer's shoes'' and think about whether someone else might answer differently. This did not seem to have any impact on results, but we noticed that smaller models seemed to express the same levels of confidence (around 90\%) in correct and incorrect or ambiguously correct decisions. gpt-4-turbo-0125 expressed nearly complete confidence in both its correct and ambiguously correct answers, but usually correctly identified alternatives to ambiguously correct answers. (An example of an ambiguously correct answer is re-tagging $a = ★b★$ as $a = ★b + c★$, since $a = ★b★ + c$ is arguably also valid.)
\end{itemize}

Even on our very small tuning set, the smaller models occasionally made serious mistakes, like moving an annotation on a function call to the function definition. They also struggled more with keeping the contents of annotations consistent at the character level: an annotation like ``for $★$line in lines:$★$'' was likely to become ``for $★$line in lines$★$:'', dropping the colon. Since breaking the problem down into smaller steps helped but did not fully eliminate these issues, there may be a fundamental difficulty for these models with the complexity and multi-step nature of the task.

The model used for our evaluation (gpt-4-turbo-0125) reliably obtained a perfect score on the tuning set with the selected prompt.

\subsection{Text point matching}

After obtaining the text and line numbers of the updated segment, our system attempts a whitespace-normalized exact match with the text in that subsection of the updated file. This naive approach gave us a perfect score on the tuning set with outputs from the model used for our evaluation. For the outputs of smaller models, we attempted to configure fuzzy matching, but did not pursue this in the system evaluated. Asking for a regular expression matching the section or the start or end of the section also did not reliably result in a correct match.

We discuss possible failures of this naive setup in section \ref{points-of-failure}.

\section{Evaluation}

We ran the re-tagging system on the benchmarks using gpt-4-turbo-0125 as the language model and report the results here.

\subsection{Results}

\subsubsection{Accuracy and points of failure}\label{points-of-failure}
79 of the 90 tags in our test set (\textasciitilde88\%) were placed with no difference at all from the ``correct'' output.

Two of the differences were incorrect matches resulting from tags beginning with whitespace, which our system was not designed to preserve. Ignoring this issue gives a ``true'' accuracy of 90\%.

Two more differences were incorrect matches due to the model stating the wrong occurence index. In our test, the model never stated any occurrence index other than 1. This seems to indicate a lack of understanding of this part of the prompt.

No match was found for seven of the benchmarks. In five of the match failures, the text was correctly identified, but the model misidentified the starting or ending line of the segment. All of these differences were off by one in the direction of starting or terminating the section of the segment early. Five of these (5, 77, 80, 82; not 20) involved what appears to be an issue with mismatched nested parentheses (see Figure \ref{fig:line-number-error-example}). Expanding the lines being searched once after a failure to match would solve this issue and give an accuracy of \~96\% on this benchmark suite. However, a new test set would be needed after making such a change. Expansion also allows incorrect matches.

\begin{figure}[H]
\begin{minted}[escapeinside=||,fontsize=\footnotesize,breaklines]{text}
4: |$★$|const PropertyListing = ({ title, address, price, bedrooms, bathrooms, image }) => {
5:  // Input validation can be implemented here if needed for additional logic
6:  return (
    ...
14:     </div>
15:   );
16: }|$★$|
\end{minted}
\caption{Code that led to an ending line number error. gpt-4-turbo-0125 chose line 15 for the final line of the segment after correctly stating the full text of the segment, including the brace on line 16.}
\label{fig:line-number-error-example}
\end{figure}

In two of the match failures, the model failed to correctly copy the text of the updated segment. One of these instances (the response for benchmark 30) omitted a comment from the updated text, and another (for 63) copied the snippet from the original file rather than the text from the updated file. 63 was the only observed misunderstanding of the primary task.

\subsubsection{Latency} Output from our prompt was about 30 tokens plus the number of tokens in the segment, which may be arbitrarily long. The average generation time for our system over the benchmarks using the gpt-4-turbo-0125 endpoint was 4.4 seconds.

\section{Discussion}

Here we discuss the implications of these results and the challenges faced by an LLM-maintained, document-external annotation system.

\subsection{Capability of current language models}

\subsubsection{Accuracy, latency, and cost tradeoffs}
Putting aside the off-by-one line number issue, gpt-4-turbo-0125's responses other than the response to benchmark 63 matched human expectations. For the kinds of re-taggings in our benchmark suite, there is no question that an existing language model is capable of the task in a vacuum, or that the accuracy could be pushed arbitrarily high by resampling.  However, the model's high cost and average response time are problematic. If cost is not a concern, since tag positions are independent, re-tagging can occur in parallel for all tags on a document, but full parallelism still requires an LLM instance for each tag. As the speed and availability of LLMs increases, this problem may diminish, but base performance will still need to increase considerably, or more creative prompts that process tags together will be needed, to scale to documents with hundreds of tags.

How else might this issue be overcome? One method we attempted was to support smaller, faster models. Mixtral-8x7B-Instruct-v0.1, a fast open source mixture-of-experts (MoE) model, usually responded in around one fifth the time of gpt-4-turbo-0125, with hosting costs around one fortieth despite performance on par with gpt-3.5-turbo-01.25. Initial tests with Mixtral were promising, but we ultimately found significant enough issues with reliability during prompt tuning (discussed in section \ref{tuning}) that we did not proceed to evaluation. This process was repeated with Gemini Pro and gpt-3.5-turbo-0125. This may indicate a fundamental lack of ability in smaller models, but more testing is needed; it may still be possible that a system with enough checks could at least solve the accuracy issue for small models, though it might require many more requests to the language model.

A more promising direction we did not explore would be to fine-tune a small model on examples collected from our code generation system (or a derived system representing a greater variety of cases). Depending on the level of success in this, fast and accurate re-tagging across many platforms could be achieved relatively soon. Another approach might use fine-tuned models for likely easier triage and validation steps to avoid calling an expensive model in most cases.

\subsubsection{Threats to validity}
As a synthesized benchmark suite, our evaluation faces obvious threats to external validity. At present, our suite represents a very low bar which any re-tagging system should clear without issue, and does not test the reliability of our system on an actual code base. At best, it establishes that gpt-4-turbo-0125 is capable of following the movement of entities in refactorings performed by the model itself. However, the failures of the smaller models reveal that this result has some value.

\subsection{What kinds of segments and edits are common but difficult to support?}

Our system and evaluation reveal at least 3 problems a semantic re-tagging system must address.

\subsubsection{``Whitespace'' and tag matching}

When building our system, we were surprised to find that the text output by the language model often differed from the input text due to whitespace. With the right prompt or model tuning this issue may go away, but without a solid solution, matching model output to the code in the buffer is a considerable challenge. Even though the system's focus is on semantics, possibly-brittle syntactic bookkeeping remains necessary.

\subsubsection{Annotation orphaning and duplication, and intent}

Our system does not attempt to handle annotations whose anchor text is simply removed from the file, and we excluded several of these cases from our benchmark. Duplicated anchor text was also excluded. A robust system must detect these cases and should probably decide whether the anchor text vanishing or multiplying means that the annotation itself should vanish or multiply. This is a more complicated problem than it may seem. For example, in an excluded benchmark, a tag on a loop condition vanished because the loop became a reduce() call. Should the tag have been re-applied to the reduce() call? Knowing that the tag referred to the $length$ of the array being looped over, maybe one would say it should not, since the reduce call has no reference to the array's length. However, if we suppose the annotation represented a comment on the loop body noting that ``the loop will execute $length$ times, so the programmer must be careful not to pass in very long arrays,'' we would probably conclude that the warning is still valid and should find a new home, perhaps on the array itself.

For these reasons our definitions in section \ref{sec:basic-definitions} include a notion of intent. However, the complexity introduced by intent led us to avoid it while testing this initial system. We hope to see an expanded set of benchmarks in the future that include full information about the content and intent of annotations in order to allow addressing cases like these.

\subsubsection{Truly ambiguous re-taggings}

Even with intent, a system will still fall short in cases where even a human would not know what the programmer wants. A robust system should detect these cases and offer suggestions for reasonable options. In limited tests on simple examples, we found gpt-4-turbo-0125 limited its responses to correct alternatives (sometimes missing alternatives), while smaller models occasionally suggested far-fetched (arguably incorrect) re-taggings.

\section{Conclusion}
In conclusion, this paper presents a novel approach to leveraging language models for the automated re-tagging and maintenance of document-external markup, highlighting the potential of LLMs to significantly enhance code annotation processes. Together, our formal description of annotations, adaptable code for benchmark generation, the synthetic Tagged Code Updates benchmark dataset, and our example implementation provide starting points for research on this technique. Our evaluation measures the viability of current language models for accurately re-tagging code in response to modifications. The gpt-4-turbo-0125 model indeed handles cases that traditional methods cannot with promising accuracy rates, but challenges such as properly matching the model's output to the file and handling annotation ambiguities remain.

Our research underscores the importance of intent and semantics in annotations and suggests a future direction for automated code maintenance tools. Ultimately, our findings lay the groundwork for further exploration into the integration of programming languages, markup systems, and artificial intelligence.

\section{Acknowledgments}
We would like to thank Kevin Shebek and Joshua Horowitz for reading early drafts of this paper and offering critical feedback. 
We also thank our reviewers for their guidance and valuable suggestions while preparing the final version of this paper.
This material is based upon work supported by the U.S. Department of Energy, Office of Science, Office of Advanced Scientific Computing Research, ComPort: Rigorous Testing Methods to Safeguard Software Porting, under Award Number DE-SC0022081.

\bibliographystyle{ACM-Reference-Format}
\bibliography{MagicMarkup}

\section{Appendix}

\subsection{Benchmark Generation}
\subsubsection{Example of a high-level problem description generated by gpt-3.5-turbo-0125.}\label{problemDescriptionExample}~\\

Problem: A local coffee shop wants to create a dynamic menu on their website that displays the prices of different coffee beverages based on the size (small, medium, large) selected by the user.

Constant: 

- menuItems: An array of objects containing information about each coffee beverage (name, small price, medium price, large price).

Steps to Solve:
\begin{enumerate}
    \item Create an array of objects called menuItems, where each object represents a different coffee beverage and stores its name, small price, medium price, and large price.
    \item Create a component that displays the menuItems in a visually appealing way, with buttons for selecting small, medium, or large sizes for each coffee beverage.
    \item Add logic to the component that updates the displayed price based on the size selected by the user.
    \item Test your component by rendering it on a webpage and verifying that the prices update correctly when different sizes are selected.
\end{enumerate}

\subsubsection{Example of a high-level snippet description for SNIPPET\_TYPE ``constant'' generated by gpt-3.5-turbo-0125.}\label{snippetDescriptionExample}~\\

One particular constant in this program is `menuItems'. It is an array of objects that represent different items on a coffee menu, each object containing the name of the item, along with the prices for small, medium, and large sizes of that item. This constant is used to display the menu items and their prices on the screen.

\subsubsection{Example of a high-level update description generated by gpt-3.5-turbo-0125.}\label{updateDescriptionExample}~\\

One potential change a real-world programmer might consider is refactoring the way the menuItems data is structured. Instead of using separate properties for smallPrice, mediumPrice, and largePrice, they might consider grouping the prices together under a single prices object for each item. This could make the data structure more flexible and easier to work with, especially if there are plans to add more size options in the future. 

In this partially applied state, the programmer may have created a new data structure for menuItems where each item now has a prices object that contains the prices for different sizes (small, medium, large). They may have updated the CoffeeMenu component to work with this new data structure for some items, but not for all items yet. As a result, some items may still be using the old separate properties for prices while others are utilizing the new prices object. 

This halfway refactored codebase may demonstrate a transitional phase where the programmer is in the process of updating the data structure and component logic to be more scalable and maintainable.

\subsubsection{Prompt to get the updated code with the snippet still properly delimited.}\label{updatePrompt}~\\

Consider the following problem:

\lstset{
  basicstyle=\ttfamily\footnotesize,
  breaklines=true,                
  breakatwhitespace=false,         
}
\begin{lstlisting}
<problemDescription>
${problemDescription}
</problemDescription>
\end{lstlisting}

Now consider this code that tries to solve the problem:

\begin{lstlisting}
<program>
${codeWithSnippetDelimited}
</program>
\end{lstlisting}

Note that a snippet from the code has been marked with a ``\$\{delimiter\}'' on both sides. This snippet is described as follows:

\begin{lstlisting}
<snippetDescription>
(${snippetType})
${snippetDescription}
</snippetDescription>
\end{lstlisting}

Now consider the following description of an update to the program:

\begin{lstlisting}
<updateDescription>
${updateDescription}
<updateDescription>
\end{lstlisting}

Apply this update to the code as described. Your response should be purely code without any external discussion, and should fully copy any relevant sections of the original program. In order to obtain credit, you MUST maintain the ``\$\{delimiter\}'' marks on the snippet or its updated version. The contents of the snippet should be functionally identical in the new version of the code.

Again, the updated version of the code MUST have a SINGLE pair of ``\$\{delimiter\}'' marks referring to the same snippet in its new position or form.

\subsection{Annotation update system}

\subsubsection{Prompt with an example program and update}\label{retag-prompt}~\\
Consider the following file:

\begin{minted}[escapeinside=||,fontsize=\footnotesize,breaklines]{text}
<INPUT>
1:
2:#include <stdio.h>
3:
4:int main() {
5:    int numItems;
6:    float totalAmount = 0;
7:    float discountedAmount = 0;
8:
9:    printf("Enter the number of items in the cart: ");
10:    scanf("%d", &numItems);
11:
12:    for (int i = 1; i <= |$★$|numItems|$★$|; i++) {
13:        float price;
14:        printf("Enter the price of item %d: ", i);
15:        scanf("%f", &price);
16:
17:        totalAmount += price;
18:    }
19:
20:    if (numItems >= 5) {
21:        discountedAmount = 0.1 * totalAmount;
22:        totalAmount -= discountedAmount;
23:    }
24:
25:    printf("\nTotal amount: $%.2f\n", totalAmount);
26:    printf("Discounted amount: $%.2f\n", discountedAmount);
27:
28:    return 0;
29:}
</INPUT>
\end{minted}

A specific segment of code has been marked with ``$★$''. The segment refers to ONLY THE TEXT BETWEEN THE ``$★$'' marks:

\begin{lstlisting}
<SEGMENT>
numItems
</SEGMENT>
\end{lstlisting}

Next, consider the following updated file:

\begin{lstlisting}
<UPDATED>
1:#include <stdio.h>
2:
3:int main() {
4:    int numItems;
5:    float totalAmount = 0;
6:    float discountedAmount = 0;
7:
8:    printf("Enter the number of items in the cart: ");
9:    scanf("%d", &numItems);
10:
11:    float prices[numItems];  // Introduce an array to store the prices of the items
12:
13:    for (int i = 1; i <= numItems; i++) {
14:        float price;
15:        printf("Enter the price of item %d: ", i);
16:        scanf("%f", &price);
17:
18:        prices[i - 1] = price;  // Store the price in the array
19:
20:        totalAmount += price;
21:    }
22:
23:    if (numItems >= 5) {
24:        discountedAmount = 0.1 * totalAmount;
25:        totalAmount -= discountedAmount;
26:    }
27:
28:    printf("\nTotal amount: $%.2f\n", totalAmount);
29:    printf("Discounted amount: $%.2f\n", discountedAmount);
30:
31:    return 0;
32:}
</UPDATED>
\end{lstlisting}

You are responsible for placing an identical annotation on this updated file. It is extremely important that you place the annotation in the correct place. Important metadata is attached to this segment.

Describe possible sections the specific segment could be said to be located in. It is possible the segment has not changed, or that it has been refactored. Pick the most correct choice. Remember to be detailed about the start and stop of the segment. If the segment has been updated, it may need to expand or shrink. BE CAREFUL TO INCLUDE NOTHING EXTRA. Then, provide the following numbered answers as a JSON object:

1) Print ONLY the text of the updated specific segment. You must print all of the text here.

2) State ONLY the line number in UPDATED that (1) starts on.

3) State ONLY the line number in UPDATED that (1) ends on.

4) (1) may occur multiple times in the section given by [(2),(3)]. Which number occurrence, as ONLY a 1-indexed number, is (1)?

The object must look like: \{1: <code>, 2: <number>, 3: <number>, 4: <number>\}

The answer to 1 should be a code string only, without markdown formatting or extra notes.

\end{document}